\documentclass[conference]{style/IEEEtran}  

\IEEEoverridecommandlockouts                              

\usepackage{graphicx}
\usepackage{epsfig} 
\usepackage{mathptmx} 
\usepackage{times} 
\usepackage{amsmath} 
\usepackage{amssymb}  
\usepackage[caption=false]{subfig}
\usepackage{nicematrix}
\usepackage{hyperref}

\title{
    Beyond the Flat Seafloor: A Closed-Form Two-View Constraint to Aid Sidescan Sonar Reconstruction
}

\author{Kalin~Norman and Joshua~G.~Mangelson%
    \thanks{This work was funded under Department of Navy awards N00014-21-1-2260 issued by the Office of Naval Research.}
  \thanks{K.~Norman and J.~Mangelson are at Brigham Young University in Provo, Utah. \texttt{\{kalinnorman, mangelson\}@byu.edu}.}
}

\begin{document}

\maketitle
\thispagestyle{empty}
\pagestyle{empty}

\begin{abstract}
Sidescan sonar is a common sensor for both manned and autonomous marine exploration and mapping, yet very few methods build upon or exploit the geometric projection model of the sensor.
As sidescan sonar is limited to a 1D range measurement, many approximations are frequently used, including the long-standing assumption of a flat seafloor.
Rather than make similar approximations, this paper focuses on a multi-view geometry based approach and formalizes a two-view geometric constraint and proves that a shared feature is constrained to a locus within the intersection of a sphere and a plane. 
In addition, we characterize what governs the size of the ambiguity locus through Monte Carlo simulation that is grounded in real aperture and mounting geometry for both a surface vessel and an underwater vehicle.
We translate additional simulations of relative trajectories for both vehicle platforms into concrete survey-planning guidance. 
Our results show that locus length is strongly governed by elevation misalignment, and peaks at a moderate oblique crossing angle of approximately 20\textdegree, with minimal locus lengths obtained at near parallel and anti-parallel passes.
\end{abstract}

\section{Introduction}

Autonomous underwater vehicles (AUVs) enable a wide range of marine exploration applications, from a greater understanding of previously inaccessible regions of the seafloor to the rapid discovery of lost or sunken objects.
Sidescan sonar (SSS) is among the most widely used sensors for these tasks, given its ability to ensonify a wide swath of seafloor at low cost relative to other sensors that provide similar coverage, such as multibeam echosounders and synthetic aperture sonar.
However, SSS has its limitations, as a single ping reduces the 3D geometry of the seafloor to a 1D range measurement, where all angular information is lost.

Reconstruction from single-modal sensor data is most developed in the optical domain, where photogrammetric structure-from-motion (SfM) recovers 3D structure from multiple 2D camera views through the geometric relationships between each view and shared features \cite{hartley2003multiple}.
Direct application of SfM algorithms to acoustic data is not possible, as in the case of SSS returns, the features are constrained to a sphere rather than a ray, and the geometric projection models differ significantly.
Consequently, the development of analogous methods to SfM for acoustic data is an open problem.

\begin{figure}[tbhp]
    \centering
    \includegraphics[width=0.7\linewidth]{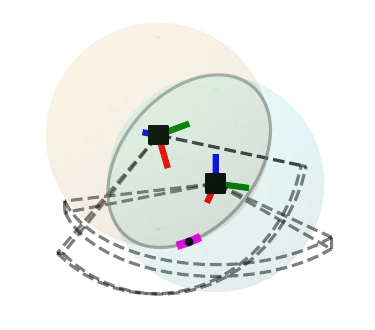}
    \caption{Visual demonstration of the geometric constraint on feature position from two sidescan sonar measurements of the same feature. Each measurement corresponds with a sphere that is centered on the sensor. The intersection of the two spheres produces a circle. Known bounds on the field-of-view from both sensors further restrict to a portion of the circle, or locus. The length of that locus is representative of the positional uncertainty on the 3D position of the feature.}
    \label{fig:locus_concept}
\end{figure}

Nevertheless, the development of reconstruction methods from sonar data is underway.
Prior work that estimated 3D structure from SSS, such as through simultaneous localization and mapping (SLAM), has approached the problem either through iterative and optimization-based methods that avoid an explicit closed-form model \cite{ruiz2004concurrent, schouten2022auv}, or by assuming a flat seafloor to convert range directly into an implicit depth \cite{cobra1992geometric, burguera2016high}.
The flat-seafloor assumption in particular is convenient, and very common, but is almost always violated in real marine operations.
Additionally, it impedes true 3D reconstruction, as the very structure that a reconstruction algorithm would need to recover is assumed known.

Building on our own recent work that formalized cross-modal (forward-looking and sidescan) acoustic stereo geometry \cite{norman2025cross}, we now focus on what can be recovered from SSS alone, using only its own projection geometry and without a flat-seafloor prior.
We show that the use of two SSS views of the same feature reduces the space of possible 3D locations not to a point (as with calibrated optical stereo) but to a 1D locus that arises from the intersection of a sphere and a plane.
This is a strictly weaker constraint than optical triangulation provides, but it is nonetheless a substantial improvement over the essentially unconstrained single-view case, and it is exact and closed-form rather than iteratively estimated.
A preliminary investigation into three (or more) view configurations shows promise for full 3D reconstruction in some cases, but a thorough treatment is beyond the scope of this work.
Instead, in this work we prioritize the two-view case.

The remainder of this paper makes three main contributions.
First, we derive the closed-form sphere-plane intersection and characterize its geometry (Sec.~\ref{sec:derivation}).
Second, we use Monte Carlo simulation, grounded in the real sensor apertures and mounting geometry of two representative platforms, to determine which aspects of relative sensor pose actually drive the size of the resulting ambiguity locus (Sec.~\ref{sec:application}).
Third, we translate this finding into concrete guidance for real surface and underwater vessel survey planning, and characterize how locus length varies with the crossing angle between two actual survey passes.

\section{Related Works}

Prior efforts toward SSS-based mapping and 3D feature point estimation are largely encompassed by a few different areas of research.
The first relies on iterative optimization to fuse SSS observations into a map or pose-graph, with the underlying 3D geometry handled implicitly through the estimator rather than through an explicit closed-form projection model.
Such methods for SSS date back at least to \cite{ruiz2004concurrent}, with related pose-graph and inertial fusion approaches for deep-sea AUV navigation \cite{woock2011side} and more recent SLAM approaches with similar formulations \cite{schouten2022auv}.
A related but distinct problem is data association within SLAM algorithms.
In an attempt to sidestep the ambiguity entirely, rather than resolve feature position, there have been approaches that either use local seafloor elevation (height) gradients \cite{mackenzie2015extracting} or matched 2D image keypoints \cite{zhang2024fully} as an association cue.
Recently, the geometry between multiple views has garnered some attention, as \cite{yang2024geometry} demonstrated how a geometry-based approach for mapping was able to outperform methods that assume a flat seafloor, while being computationally efficient.

Other approaches invert a Lambertian model of the SSS backscatter, or acoustic shadow, to recover the shape of observed structures.
This is commonly referred to as shape-from-shading (SfS) and is typically applied to a single survey pass \cite{coiras2007multiresolution, bikonis2013application, moszynski2013reconstruction}.
More recent variants have fused SfS with learned monocular depth estimates \cite{ju2024three} or self-consistency constraints \cite{zhao2018reconstructing} to improve estimation accuracy.
While similar to our approach in that they avoid the assumption of a flat seafloor, these methods recover shape through the inversion of an acoustic model, and are generally restricted to a single pass, rather than through the use of multiple sensor observations.

A more recent body of research fuses many SSS pings across a full survey to estimate bathymetry through the use of deep neural networks, again without assuming a flat seafloor.
This body of work contains the closest conceptual neighbors to our work, as they target multi-view SSS reconstruction without the assumption of a flat seafloor.
One of the more direct approaches is \cite{xie2022sidescan} that removes the need for external altimeter data via a nadir-region geometric constraint.
Some of the works that followed include \cite{xie2022neural, xie2025neurss}, which add neural surface-normal estimation and joint SLAM optimization, respectively, and \cite{bore2022neural}, which instead enforces self-consistency across many overlapping lines via an implicit neural SfS model.
Notwithstanding the similar goal, these approaches differ significantly from our approach in their reliance on a globally-optimized learned model rather than a closed-form geometric constraint.

Finally, since some of the earliest SSS image-processing pipelines \cite{cobra1992geometric}, the assumption of a flat seafloor has dominated most SSS uses.
This assumption converts an SSS range return almost directly into a 3D point by intersecting the range with a flat plane determined by the height from bottom of the vehicle.
However, this assumption is violated by nearly all seafloor environments, which is a limitation whose practical impact has been directly quantified \cite{burguera2016high}.
Consequently, as demonstrated by the more recent literature, many works have already begun to move away from the assumption, and we continue that trend.
Instead, our focus is on the exploitation of known geometric information from multiple views.

Our own prior work \cite{norman2025cross} formalized closed-form stereo projection geometry between a sidescan and a forward-looking sonar.
This paper extends that geometric approach to the single-modality two-view SSS case and, to the best of our knowledge, is the first to characterize the resulting ambiguity purely from acoustic projection geometry.
We intentionally do so without an iterative estimator, a learned model, or a flat-seafloor prior.

\section{Preliminaries}
\label{sec:preliminaries}

\subsection{Sensor Model}

An SSS emits an acoustic ping at a known frequency (or range of frequencies) and measures the round-trip time of flight of any returns, converted via the local speed of sound into range measurements.
We adopt a local sensor coordinate frame, as in \cite{norman2025cross}, with the sensor oriented in the $+x$ direction, with $+y$ to the left of the sensor, and $+z$ up, and describe the location of a feature relative to the sensor in spherical coordinates $(r, \theta, \phi)$, the range, azimuth, and elevation, which are related to local Cartesian coordinates through
\begin{equation}
    \begin{bmatrix} 
        x \\ 
        y \\ 
        z 
    \end{bmatrix} = \begin{bmatrix} 
    r \cos \theta \cos \phi \\ 
    r \sin \theta \cos \phi \\
    r \sin \phi
    \end{bmatrix} .
    \label{eq:spherical}
\end{equation}

A physical SSS transducer has finite azimuth and elevation apertures, $|\theta|\le\theta_{\max}$ and $|\phi|\le\phi_{\max}$.
In addition, SSS apertures are characteristically very wide in azimuth ($\theta_{\max}$ on the order of tens of degrees) and extremely narrow in elevation ($\phi_{\max}$ often around or under one degree).
To relate our local frame to other common terms for sidescan operations, the azimuth corresponds to the across-track aperture, which must span the full imaged swath, while elevation corresponds to the along-track beam width, which sets along-track resolution as the vehicle travels.

\subsection{Vehicle Mounting}

\begin{figure}[t]
    \centering
    \subfloat[Boat, top-down]{
        \includegraphics[width=0.45\linewidth]{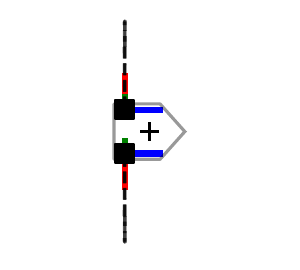}
    }
    \subfloat[Boat, stern]{
        \includegraphics[width=0.45\linewidth]{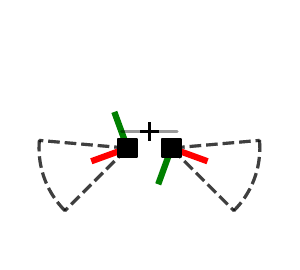}
    } \\
    \subfloat[AUV, top-down]{
        \includegraphics[width=0.45\linewidth]{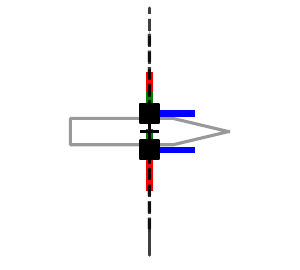}
    }
    \subfloat[AUV, stern]{
        \includegraphics[width=0.45\linewidth]{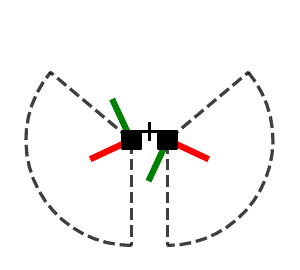}
    }
    \caption{Local coordinate frame ($x$=red, $y$=green, $z$=blue) and frustum for the mounted transducers. The top row shows the surface vessel setup with dual Omniscan 450 SS, and the bottom row the AUV with dual EdgeTech 2205. Each vessel is shown from two perspectives, with the left column showing top-down and the right column from the stern looking toward the bow. The plotted maximum range is chosen per platform for legibility of the mounting geometry and not the real operating range of the sensors.}
    \label{fig:rig}
\end{figure}

A common configuration mounts two transducers on a single vehicle, one to port and one to starboard, each tilted downward and outward from the vehicle body so that the narrow elevation aperture sweeps roughly along the direction of travel while the wide azimuth aperture sweeps below and to the sides of the vehicle.
We ground our simulations in two platforms, one intended to represent a surface vehicle and the other an AUV.
Given the popularity and accessibility of the BlueRobotics BlueBoat, our surface vessel is based on the measurements of that vehicle, and the SSS used in our simulation is based on the Omniscan 450 SS, which is a common pairing for the BlueBoat.
For the underwater vehicle, we base the SSS mounting dimensions on a torpedo-shaped submersible, such as an OceanServer IVER or REMUS, equipped with an approximation of an EdgeTech 2205 SSS system.
Table~\ref{tab:sensors} summarizes the aperture and range specifications used throughout; mounting tilt and offsets are taken from manufacturer documentation where published and otherwise estimated from common vehicle dimensions.
Fig.~\ref{fig:rig} shows the local coordinate frame convention, the resulting frustum, and the mounted rig layout for both platforms.

\begin{table}[t]
\centering
\caption{Sonar configurations used in simulation.}
\label{tab:sensors}
\begin{tabular}{lccc}
\hline
 & Azimuth ap. & Elevation ap. & Max range \\
\hline
Omniscan 450 SS & 50\textdegree & 0.5\textdegree & 150 m \\
EdgeTech 2205 & 130\textdegree & 0.27\textdegree & 150 m \\
\hline
\end{tabular}
\end{table}

\subsection{Assumptions}

Throughout the closed-form derivation of Sec.~\ref{sec:derivation}, we assume known data association (the same physical feature is correctly matched between the two views), as well as known relative pose between the two sensor observations.
In Sec.~\ref{sec:application} we retain both assumptions but explore how the resulting ambiguity varies across the full space of relative poses consistent with the two platforms above, rather than assuming any single baseline is representative.

\section{Closed-Form Geometric Constraint}
\label{sec:derivation}

Consider two SSS observations of the same feature from two distinct sensor poses, either from two separate measurements from the same vehicle, or observations from two different vehicles.
Let $\mathbf{p}_1 = [x_1\ y_1\ z_1]^T$ be the unknown location of the feature in the local frame of the first sensor.
In addition, let the known rigid transformation from the first to the second sensor frame be given by rotation $\mathbf{R}$ and translation $\mathbf{t}$, such that the location of the feature in the frame of the second sensor is
\begin{equation}
    \mathbf{p}_2 = \mathbf{R} \mathbf{p}_1 + \mathbf{t}.
    \label{eq:transform}
\end{equation}
Each sensor directly measures only the range to the feature, and that range is related to the location of the feature through
\begin{equation}
    \|\mathbf{p}_1\| = r_1, \qquad \|\mathbf{p}_2\| = r_2 .
    \label{eq:ranges}
\end{equation}
Consequently, $\mathbf{p}_1$ is constrained to a sphere of radius $r_1$ centered at the first sensor, and $\mathbf{p}_2$ to a sphere of radius $r_2$ centered at the second.
This constraint differs significantly from optical stereo configurations, where each camera measurement produces a ray, and the reconstructed point lies at the intersection of the two rays.
For SSS, the two spheres associated with each range measurement intersect in a circle.

We substitute \eqref{eq:transform} into $\|\mathbf{p}_2\|^2=r_2^2$ and expand to get
\begin{align}
    ( \mathbf{R} \mathbf{p}_1 + \mathbf{t} )^T ( \mathbf{R} \mathbf{p}_1 + \mathbf{t} ) &= r_2^2 \notag \\
    \| \mathbf{p}_1 \|^2 + 2 \mathbf{t}^T \mathbf{R} \mathbf{p}_1 + \| \mathbf{t} \|^2 &= r_2^2 ,
\end{align}
using $\mathbf{R}^T \mathbf{R} = \mathbf{I}$.
Next, substitute $\| \mathbf{p}_1 \|^2 = r_1^2$ from \eqref{eq:ranges} to get
\begin{align}
    r_1^2 + 2 \mathbf{t}^T \mathbf{R} \mathbf{p}_1 + \| \mathbf{t} \|^2 &= r_2^2 \notag \\
    2 \mathbf{t}^T \mathbf{R} \mathbf{p}_1 &= r_2^2 - r_1^2 - \| \mathbf{t} \|^2
\end{align}
which leads to a simplified linear constraint in the form of
\begin{equation}
    2 \mathbf{u}^T \mathbf{p}_1 = C, \qquad \mathbf{u} \triangleq \mathbf{R}^T \mathbf{t}, \qquad C \triangleq r_2^2 - r_1^2 - \| \mathbf{t} \|^2,
    \label{eq:plane}
\end{equation}
with $\mathbf{u}$ and $C$ both known given a known relative pose and the two range measurements.

Equation~\eqref{eq:plane} is the equation of a plane.
Combined with the sphere constraint $\| \mathbf{p}_1 \| = r_1$, this proves that two-view SSS geometry constrains the feature to the intersection of a sphere and a plane, which produces a circle rather than a point.

Parameterizing $\mathbf{p}_1$ in spherical coordinates via \eqref{eq:spherical} and substituting into \eqref{eq:plane} gives
\begin{equation}
    2 r_1 ( u_x \cos \theta_1 \cos \phi_1 + u_y \sin \theta_1 \cos \phi_1 + u_z \sin \phi_1 ) = C,
    \label{eq:onelocus}
\end{equation}
noting that $\mathbf{u} = \begin{bmatrix}
    u_x & u_y & u_z
\end{bmatrix}^T$.
This result is a single equation with two unknown angles $\theta_1,\phi_1$.
If either angle were independently known, \eqref{eq:onelocus} would reduce to a directly solvable closed-form expression.
However, without additional information the system is under-constrained by exactly one degree of freedom, and the solution is bounded by a 1D locus produced by a sphere-plane intersection.

\begin{figure*}[t]
    \centering
    \subfloat[]{
        \includegraphics[width=0.3\linewidth]{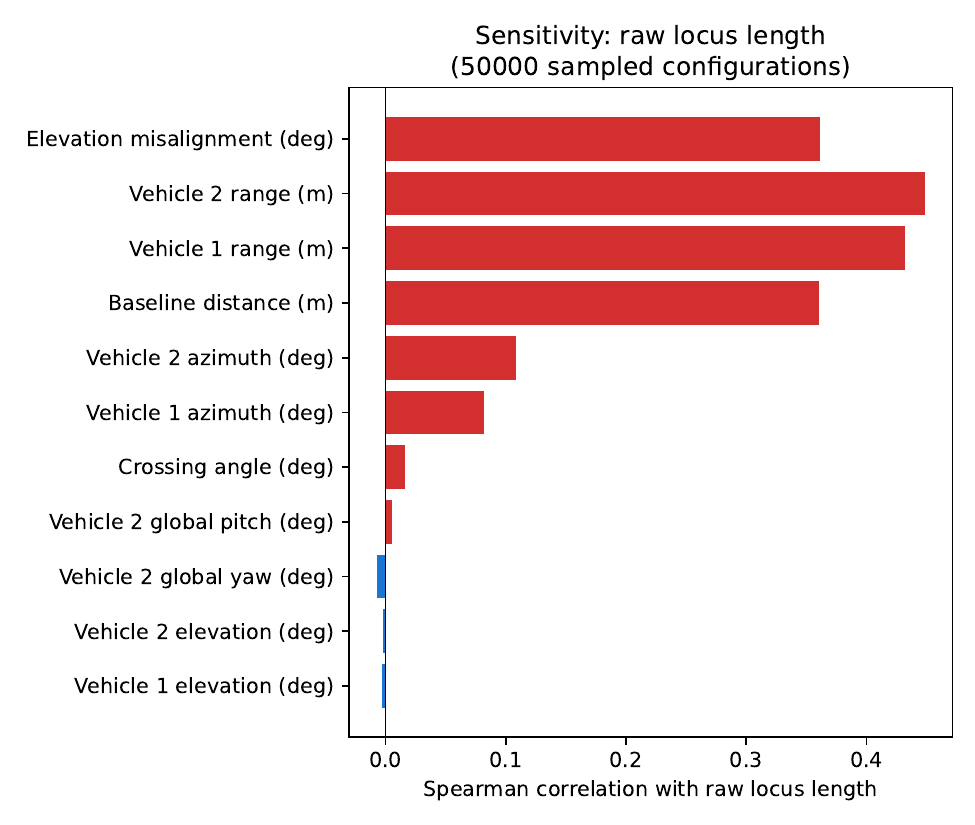}
    }
    \subfloat[]{
        \includegraphics[width=0.3\linewidth]{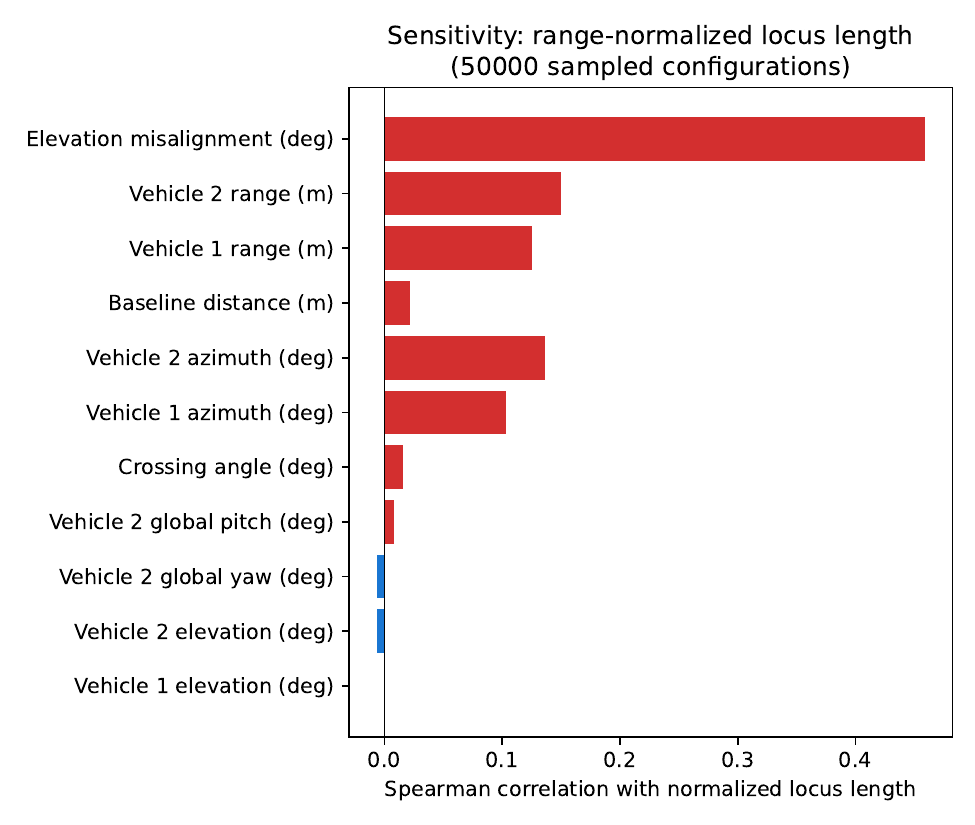}
    } 
    \subfloat[]{
        \includegraphics[width=0.3\linewidth]{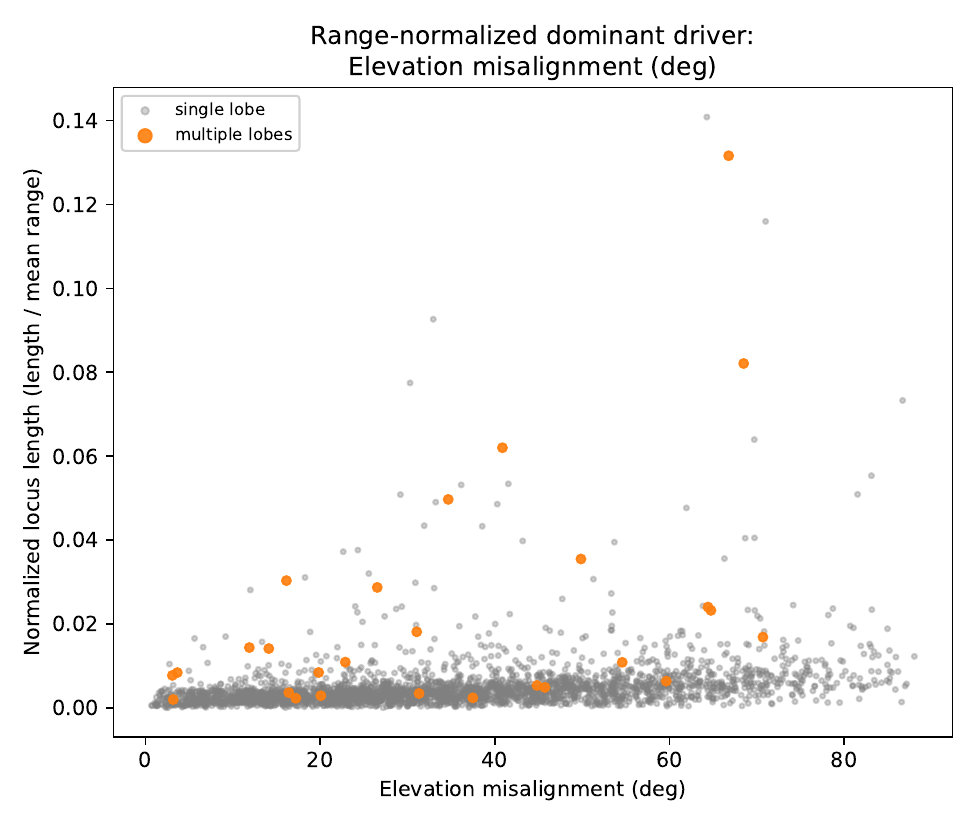}
    }
    \caption{Sensitivity of the ambiguity locus to candidate relative-pose quantities for a dual AUV configuration. (a, b) Spearman correlation with raw locus length and with locus length normalized by mean range, respectively. Range and baseline distance dominate the raw ranking but drop sharply once range is normalized out, leaving elevation misalignment as the dominant driver, with local azimuth angles also exhibiting a noticeable correlation. (c) Range-normalized locus length vs. elevation misalignment for the same samples (subsampled for legibility). Orange points mark the minority of configurations where the ambiguity locus fragments into multiple disjoint arcs rather than one contiguous arc.}
    \label{fig:sensitivity}
\end{figure*}

Due to the restricted field-of-view (FOV) of each sensor, this locus is not observed in its entirety.
The finite azimuth and elevation apertures (Sec.~\ref{sec:preliminaries}) crop the circle to the portion that falls within the FOV of both sensors.
In particular for SSS the elevation aperture is typically extremely narrow and the resulting visible arc length, which we call the locus length for the remainder of this paper, is a small fraction of the full circumference of the circle.
It is this locus length, which is visualized in Fig. \ref{fig:locus_concept}, that we investigate through simulation in the sections that follow, as it is the quantity that governs the practical positional uncertainty for any attempted reconstruction of the feature locations from two SSS views.

\section{Application and Analysis}
\label{sec:application}

\subsection{Simulation Methodology}

Initially, we verified our approach on an idealized, unconstrained sensor pair and conducted a full-range Monte Carlo simulation over relative position and orientation.
However, we determined that this idealized sweep was not representative, as many of the sample relative orientations corresponded with vehicle attitudes that are either physically impossible or operationally impractical, such as an observation that comes from beneath the seafloor, or a vehicle rotated to be on its side.
We therefore grounded the simulation in our two representative real platforms (Sec.~\ref{sec:preliminaries}), and constrained relative orientation to what each vehicle would likely experience or achieve, and repeated the analysis.
For the surface vessel we restricted it to have roll$=$pitch$=0$\textdegree, reflecting level operation in calm water, and roll$=0$\textdegree\ with pitch sampled within $\pm45$\textdegree\ for the AUV, to better represent the attitude range achievable by a torpedo-shaped AUV.
All of the results that follow use this vehicle-grounded approach, evaluated with the closed-form geometric model of Sec.~\ref{sec:derivation} cropped to the real aperture of each sensor (Table~\ref{tab:sensors}).
Because only the relative configuration between the two views matters, we fixed the position/orientation of the first sensor without loss of generality and sampled the position of the feature within the aperture and range of the sensor.
We then independently sampled the range, aperture position, and orientation of the second sensor, within the aforementioned vehicle limits.

\subsection{What Drives Locus Length}

We first investigate which aspects of the relative pose actually predict locus length.
To determine this we used the data from our Monte Carlo simulation of the two vehicle poses and ranked candidate quantities by their Spearman rank correlation with locus length.

One of the most predictable influences on locus length is the range from the sensor to the feature: if both ranges and the baseline between sensors are scaled uniformly by a factor $k$, holding all angles fixed, the resulting locus length scales by exactly $k$ as well.
To limit the influence range has on determining other major factors, we also conducted the same series of tests but with locus length normalized by range.
This enabled the discovery of a more unexpected driver of the locus length, that we refer to as the elevation misalignment.
Elevation misalignment is determined by taking the maximum of two independent elevation angles, one for each local frame of the two sensors.
The angle is the local elevation angle of the ray that intersects the two sensor origins, and we denote the angles with $\psi_1$ and $\psi_2$.
After calculating those two elevation angles, the elevation misalignment is given by the function $\max(\psi_1, \psi_2)$.
Fig.~\ref{fig:sensitivity} shows the ranking for both the non-normalized and range-normalized cases, alongside the relationship between elevation misalignment and locus length.

\subsection{Where It Matters Most}

\begin{figure*}[t]
    \centering
    \subfloat[]{
        \includegraphics[width=0.3\linewidth]{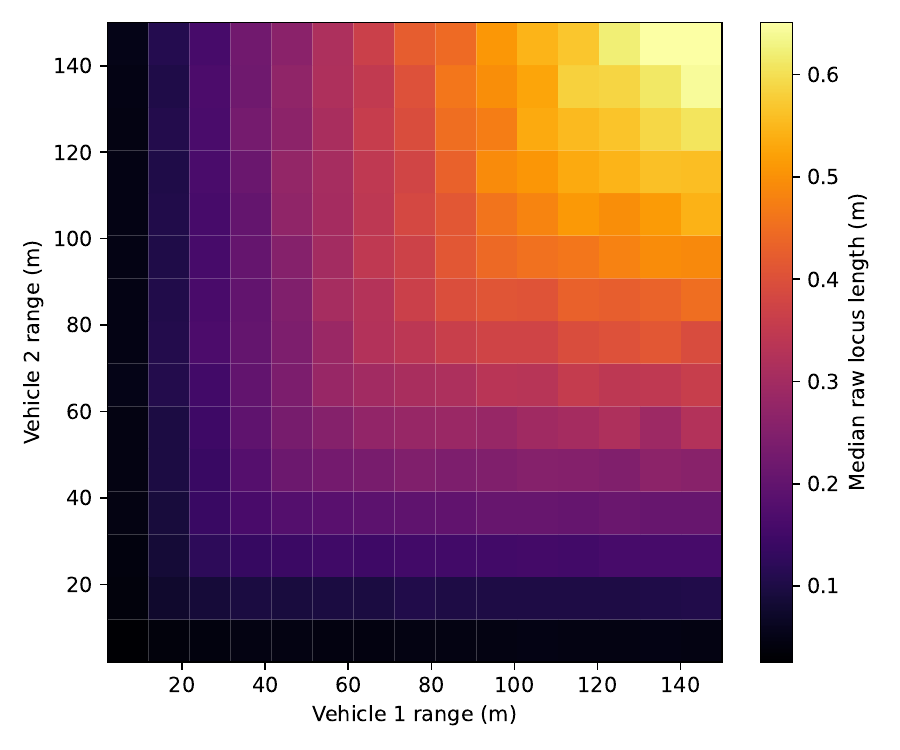}
    }
    \subfloat[]{
        \includegraphics[width=0.3\linewidth]{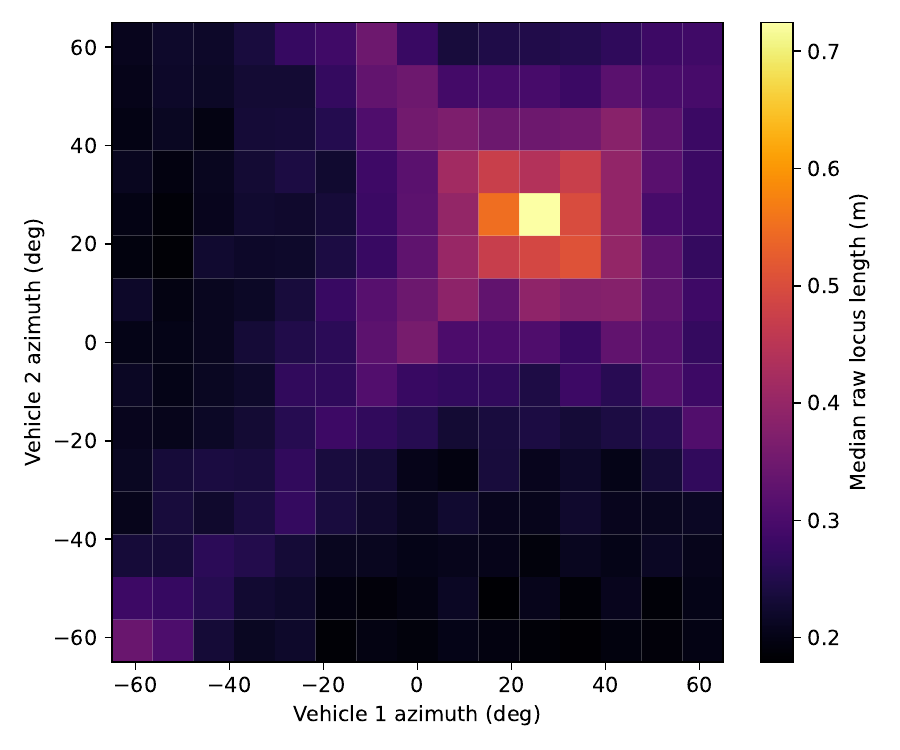}
    } 
    \subfloat[]{
        \includegraphics[width=0.3\linewidth]{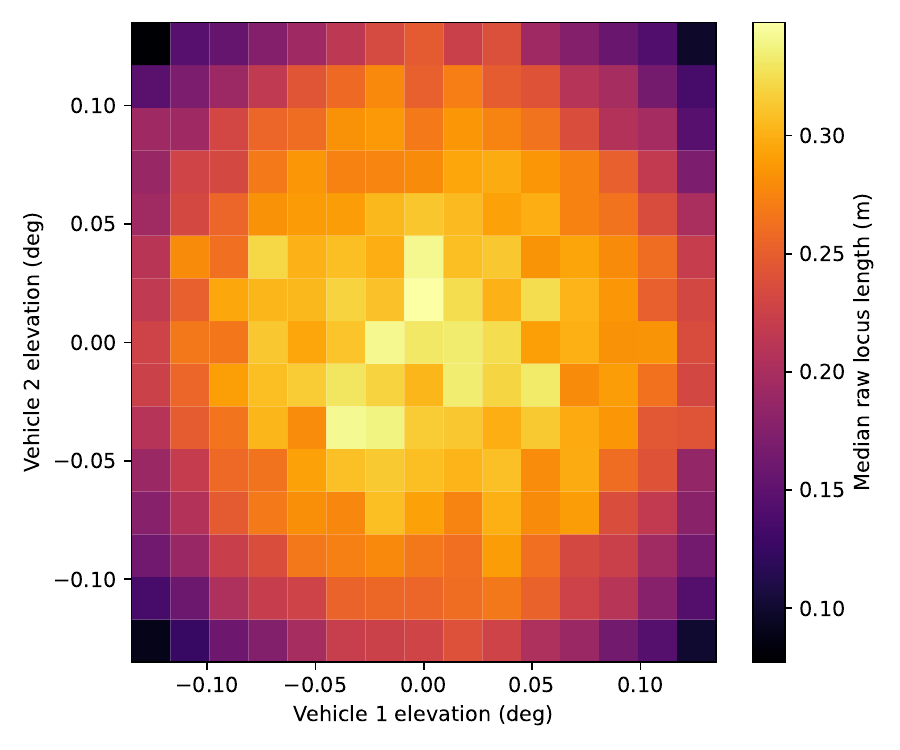}
    } \\
    \subfloat[]{
        \includegraphics[width=0.3\linewidth]{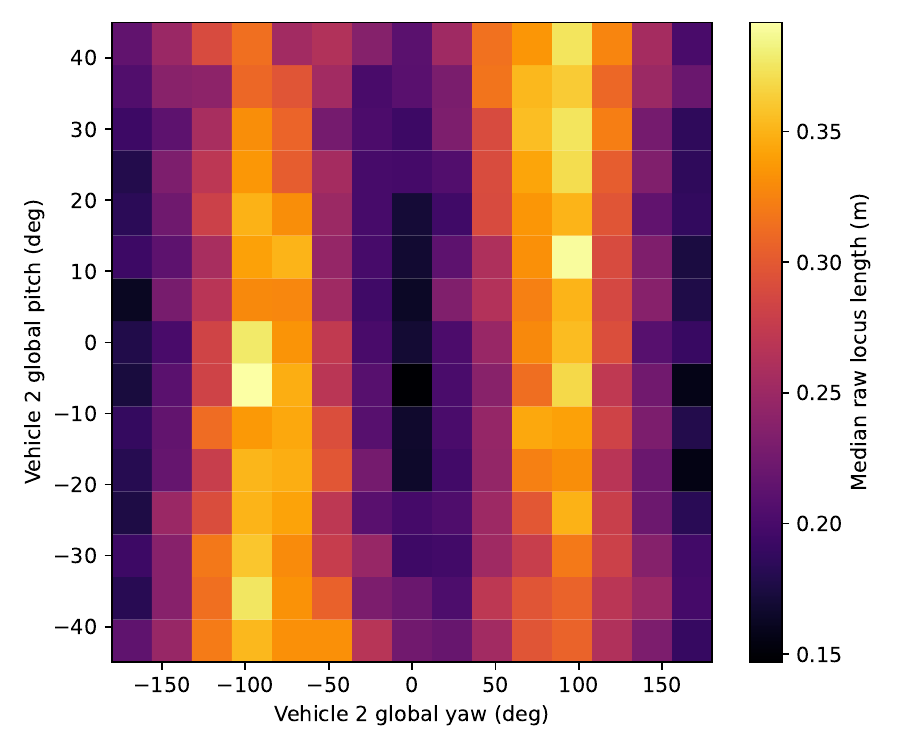}
    } 
    \subfloat[]{
        \includegraphics[width=0.3\linewidth]{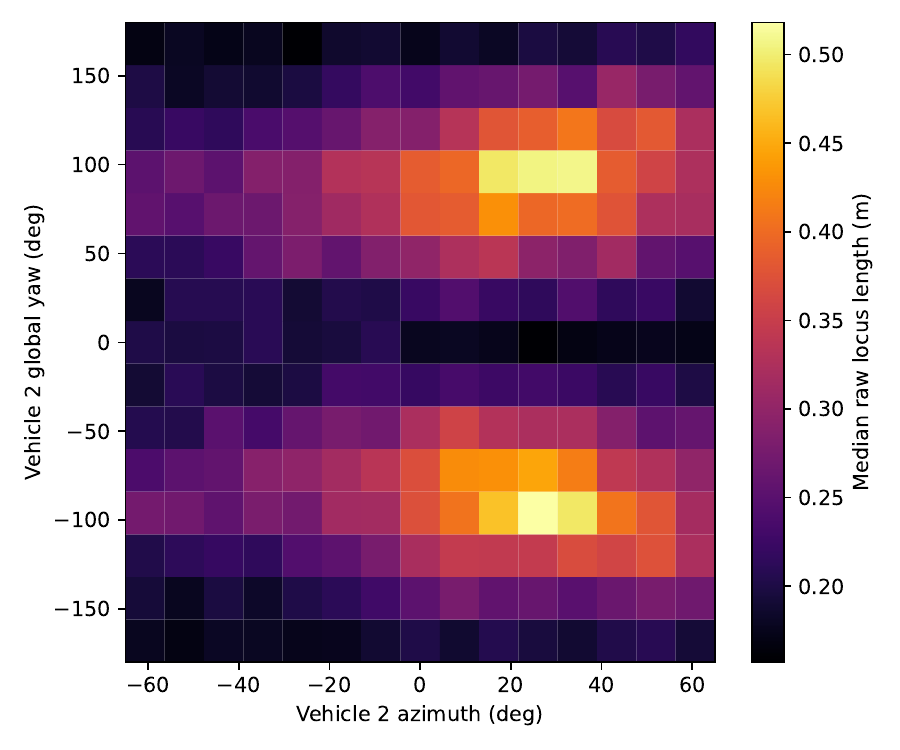}
    }
    \caption{Non-normalized locus length across pairs of relative-pose quantities with the feature always identified on the port side of the vehicle, marginalizing over the remaining dimensions, with the median value shown per bin. (a) Range vs. range: locus length grows with the range of both vehicles. (b) Azimuth vs. azimuth: a unique hotspot when both views put the feature around positive $20$\textdegree--$40$\textdegree. (c) Elevation vs. elevation: resembles a bivariate Gaussian bump centered at $0$. (d) The global yaw of vehicle 2 vs. pitch (vehicle 1 fixed at $0$ for both as a reference): a clear, roughly symmetric, double-peaked structure. (e) The azimuth of vehicle 2 vs. its own global yaw: once again two main peaks along the yaw axis, though now affected by the azimuth, isolating the hotspots more than they appear in (d).
}
    \label{fig:heatmaps}
\end{figure*}

The ranking in Fig.~\ref{fig:sensitivity} establishes which quantity matters most on average, but a rank correlation alone cannot reveal non-monotonic structure.
Fig.~\ref{fig:heatmaps} shows non-normalized locus length as a function of pairs of relative-pose quantities, marginalizing over the rest, noting that the feature was always on the port side of the vehicle.
Given the symmetry of the system, the starboard side results can be understood from the port side results we include in this document.

The total number of parameter combinations far exceeds what we can reasonably show, so we isolated a few of the most interesting or useful quantities that we could identify.
Range behaves as expected, with locus length growing as the range of both sensors increases.
Azimuth position within the aperture of each sensor shows a real, non-trivial hotspot rather than a simple monotonic trend.
Elevation is rather minimal in its effect, though it is noteworthy that it is not uniform, but rather closely resembles a bivariate Gaussian bump centered near zero misalignment in the elevation of both sensors.
Most notably, pairings that include the global yaw of the second sensor demonstrate a clear, and roughly symmetric, double-peaked structure.
Other parameters do impact the peaks of that structure, as can be seen in the azimuth and yaw pairing.
In general, we observe a large locus length across a fairly substantial range of the relative yaw, with the smallest locus lengths found near yaw~$\approx0$\textdegree\ and near the extremes of yaw~$\approx \pm 180$\textdegree, which indicate that parallel and anti-parallel observations of the feature help reduce the positional ambiguity.

\subsection{From Pose Space to Survey Trajectories}

The pose-space results characterize instantaneous relative-pose sensitivity, but the practically relevant question for mission planning is how the choice of survey trajectory affects locus length.
We simulate a vehicle running two straight survey passes over a feature at a fixed location, and vary the crossing angle between the two passes from $0$\textdegree\ (an exact repeat pass) through $90$\textdegree\ (a perpendicular cross-line) to $180$\textdegree\ (a fully anti-parallel pass), for both the surface vessel and AUV vehicle pairings.
The surface vessel remains level and at a fixed altitude throughout, as dictated by its physical operation at the surface, while the AUV can operate at variable depths and pitches between each pass.
For these experiments we allowed pitch variation to be within a realistic $\pm15$\textdegree\ range, which we note is narrower than the $\pm45$\textdegree\ envelope used in Sec.~\ref{sec:application}.
The prior envelope was meant to bound what an AUV could ever physically achieve rather than what it would realistically do mid-survey, which is why for these experiments we restricted it to a narrower range.

\begin{figure}[t]
    \centering
    \subfloat[]{
        \includegraphics[width=0.7\linewidth]{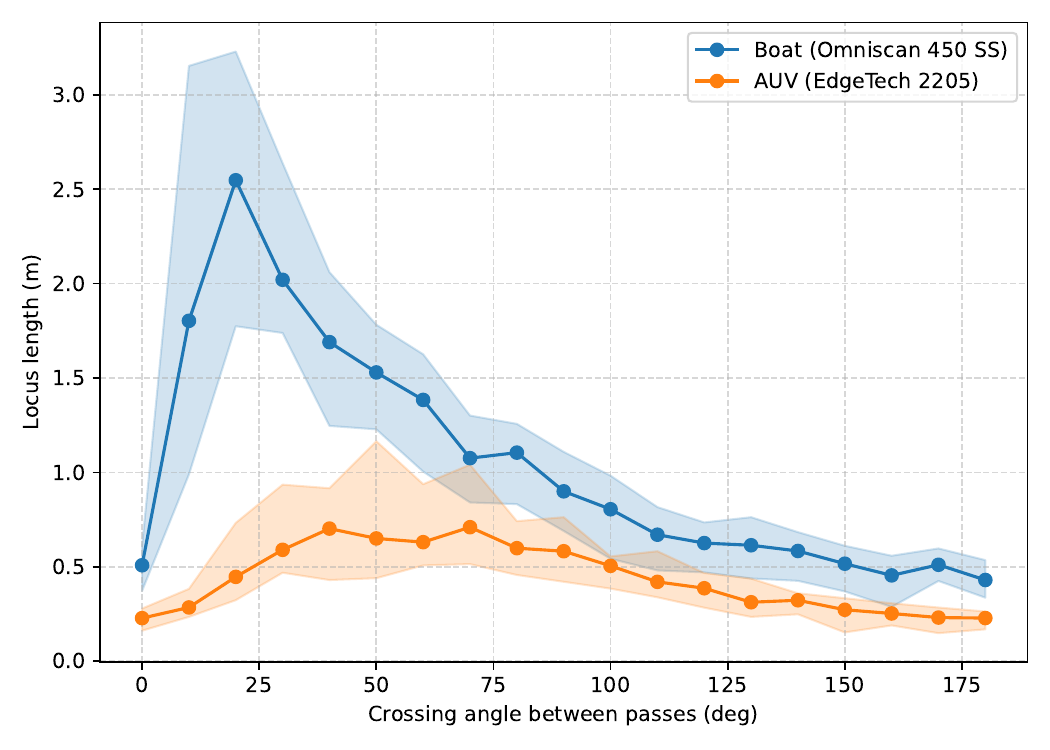}
    } \\
    \subfloat[]{
        \includegraphics[width=0.7\linewidth]{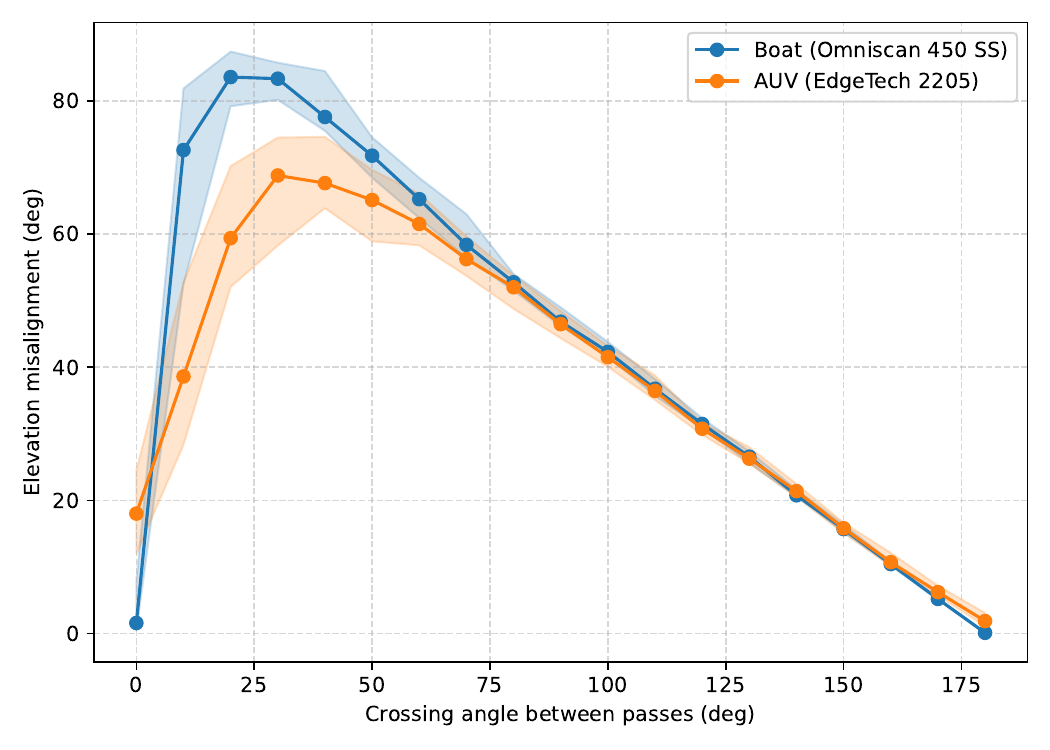}
    } 
    \caption{(a) Ambiguity locus length and (b) elevation misalignment as a function of the crossing angle between two survey passes for both the surface vessel with dual Omniscan 450 SS and AUV with dual EdgeTech 2205 platforms, each approximating the altitude/pitch limitations per-vehicle. The median and inter-quartile range are shown over repeated sampling of a representative operating range. Both platforms show a non-monotonic relationship, peaking at a moderate oblique crossing angle ($\approx20$\textdegree) rather than at a repeated pass ($0$\textdegree) or a fully anti-parallel trajectory ($180$\textdegree). The lower absolute locus length for the AUV reflects its narrower elevation aperture, and the ability to operate at both at variable heights and pitch angles between passes, while the surface vessel is fixed at sea level. Even with the differences in vehicular capabilities, the qualitative crossing-angle trend is similar for both platforms.}
    \label{fig:trajectory}
\end{figure}

The result (Fig.~\ref{fig:trajectory}) is a non-monotonic dependence on crossing angle.
Locus length rises sharply from a repeat pass to a pronounced peak at a moderate oblique crossing angle near $20$\textdegree, then declines steadily as the crossing angle increases further, continuing to improve through a perpendicular crossing ($90$\textdegree) and reaching its lowest values near a fully anti-parallel pass ($180$\textdegree).
A near-repeat pass ($0$\textdegree) performs comparably well to the anti-parallel extreme, so the two safest choices are a close-to-repeat pass or a close-to-anti-parallel pass, while a perpendicular crossing is a reasonable, though sub-optimal, middle ground.
This tracks elevation misalignment almost exactly, and both quantities show the same shape for both platforms.
The result is, at first glance, counter to classical stereo/triangulation intuition from optical setups, which treats a near-zero baseline as the worst possible case \cite{hartley2003multiple}.
For SSS, a repeat pass does produce an enormous uncropped ambiguity circle, since two nearly-identical range measurements barely constrain anything. 
Nevertheless, the FOVs of the sensors crop the locus and counter this effect, and the locus will be very short so long as the elevation misalignment remains low.
Near-repeat passes do have a very low elevation misalignment, as both passes share the same heading and hence the same elevation-zero plane.

In summary, our main operational recommendation to minimize positional ambiguity is to plan survey passes that avoid oblique crossing angles in roughly the $10$\textdegree--$40$\textdegree range, and instead to prefer either a near-repeat pass or anti-parallel pass.
We note that a perpendicular crossing is not optimal, but does represent a substantial improvement over the worst-case oblique angles and remains a reasonable fallback where a repeat or anti-parallel pass is impractical.
The operational guidance we provide does not yet take into account the effects of measurement noise and positional uncertainty in the relative pose. Due to limited space we have left those as areas of future work.

For the AUV specifically, we also examined how pitch within a range of $\pm15$\textdegree\ affects locus length, regardless of the crossing angle.
Within the worst crossing-angle band identified above, a nose-up attitude increased the median locus length by roughly $40$\%, relative to a nose-down attitude of the same magnitude, from which we conclude that a slight nose-down trim is preferable to nose-up during a survey pass.
We found no comparably clear effect from the choice of operating depth of the AUV.

\section{Conclusion}

This paper set out to characterize what multi-view SSS geometry alone can say about the 3D location of an observed feature, without a flat-seafloor prior, iterative estimator, or learned model.
We showed that two SSS range views constrain a feature to the intersection of a sphere and a plane, a 1D locus rather than a point.
Through Monte Carlo simulation grounded in real vehicle and sensor geometry, we showed that the practical size of the locus is governed primarily by both range from the sensor and elevation misalignment.
To translate this into concrete survey-planning terms for both platforms studied, based upon our simulation results, a mission should avoid oblique crossing angles between passes in roughly the $10$\textdegree--$40$\textdegree range, preferring instead a near-repeat pass or a fully anti-parallel pass.
In particular for AUVs, maintaining a slight nose-down pitch during a pass further reduces positional ambiguity, though we found no comparable benefit from the choice of operating depth.
A natural next step, left to future work, is to determine under what conditions a third (or higher-order) view collapses this locus to a fully-resolved 3D point, which would establish a complete geometry-based acoustic SfM pipeline.
Additionally, extending the present noise-free geometric analysis with a formal measurement-uncertainty model, as well as the use of the pose-space sensitivity results to directly inform automated survey path planning are further directions to pursue.
We hope this geometric characterization contributes toward mapping methods better suited to the unstructured, almost never flat, ocean floor.

\section*{Acknowledgments}

The authors acknowledge the use of Anthropic's Claude in the generation and refinement of code for the figures included in our analysis, as well as for textual refinement and drafting of this manuscript.
All underlying research questions, mathematical formulations, and interpretive conclusions are the original work of the authors, and all generated code has been reviewed and refined by the authors.

\bibliographystyle{references/IEEEtran}
\bibliography{references/IEEEabrv, references/references}

\end{document}